\documentclass[lettersize,journal]{IEEEtran}
\usepackage{amsmath,amsfonts}
\usepackage{algorithmic}
\usepackage{algorithm}
\usepackage{array}
\usepackage[caption=false,font=normalsize,labelfont=sf,textfont=sf]{subfig}
\usepackage{textcomp}
\usepackage{stfloats}
\usepackage{url}
\usepackage{verbatim}
\usepackage{graphicx}
\usepackage{cite}
\usepackage{graphicx}
\usepackage{amsmath}
\usepackage{amssymb}
\usepackage{booktabs}
\usepackage{multirow}
\usepackage{algorithm, algorithmic}
\usepackage{bbm}
\usepackage{soul, color, xcolor}
\usepackage{ulem}
\usepackage[switch]{lineno}

\usepackage[capitalize]{cleveref}
\crefname{section}{Sec.}{Secs.}
\Crefname{table}{Table}{Tables}
\Crefname{equation}{Eq.}{Eqs.}

\begin{document}

\title{Generalized Face Liveness Detection via De-fake Face Generator}

\author{Xingming Long,~\IEEEmembership{Student Member,~IEEE,},
        Jie Zhang,~\IEEEmembership{Member,~IEEE,},
        Shiguang Shan,~\IEEEmembership{Fellow,~IEEE,}
                
\thanks{This work is partially supported by National Key R\&D
Program of China (No. 2021YFC3310100), Strategic Priority Research Program of the Chinese Academy of Sciences (No. XDB0680202),
Beijing Nova Program (20230484368), and Youth Innovation Promotion Association CAS.}
\thanks{This work involved human subjects or animals in its research. Approval of all ethical and experimental procedures and protocols was granted by Ethics Committee of Institute of Computing Technology, Chinese Academy of Sciences.}
\thanks{Xingming Long, Jie Zhang and Shiguang Shan are with the Key Laboratory of AI Safety of CAS, Institute of Computing Technology (ICT), Chinese Academy of Sciences (CAS), Beijing 100190, China, and also with the University of Chinese Academy of Sciences (UCAS), Beijing 100049, China (e-mail: xingming.long@vipl.ict.ac.cn; zhangjie@ict.ac.cn; sgshan@ict.ac.cn).}
\thanks{\textit{(Corresponding author: Jie Zhang)}}
}

\markboth{Journal of \LaTeX\ Class Files,~Vol.~14, No.~8, August~2021}%
{Shell \MakeLowercase{\textit{et al.}}: A Sample Article Using IEEEtran.cls for IEEE Journals}


\maketitle

\begin{abstract}
Previous Face Anti-spoofing (FAS) methods face the challenge of generalizing to unseen domains, mainly because most existing FAS datasets are relatively small and lack data diversity. Thanks to the development of face recognition in the past decade, numerous real face images are available publicly, which are however neglected previously by the existing literature. 
In this paper, we propose an Anomalous cue Guided FAS (AG-FAS) method, which can effectively leverage large-scale additional real faces for improving model generalization via a De-fake Face Generator (DFG). Specifically, by training on a large-scale real face only dataset, the generator obtains the knowledge of what a real face should be like, and thus has the capability of generating a “real” version of any input face image. 
Consequently, the difference between the input face and the generated “real” face can be treated as cues of attention for the fake feature learning. With the above ideas, an Off-real Attention Network (OA-Net) is proposed which allocates its attention to the spoof region of the input according to the anomalous cue.
Extensive experiments on a total of nine public datasets show our method achieves state-of-the-art results under cross-domain evaluations with unseen scenarios and unknown presentation attacks. Besides, we provide theoretical analysis demonstrating the effectiveness of the proposed anomalous cues.
\end{abstract}


\section{Introduction}\label{sec1}

\IEEEPARstart{I}{n} today's world, face recognition systems are extensively utilized for identity authentication in various scenarios. However, these systems are susceptible to presentation attacks, such as using a video or a photograph of a person to deceive the system, thereby posing a significant security threat.
In order to address the security threat, significant efforts have been made in the field of Face Anti-Spoofing (FAS), also known as Presentation Attack Detection (PAD) according to the ISO definition~\cite{ISO_2023}, to defend against potential presentation attacks on face recognition systems.
Researchers have constructed a diverse set of Face Anti-spoofing (FAS) datasets~\cite{CASIA2012, MSU2015, REPLAY2012, OULU2017, CelebA-Spoof2020, HKBU2016, WFFD2020, SiW2018, Unsupervised_DA_Rose-Youtu2018, WMCA2019} and investigated the possibility of conducting face presentation attack detection on these datasets.
Previous works have achieved excellent performance on each of these datasets under intra-dataset evaluation~\cite{LBP_HoG2012, LBP2015, Color2016, distortion2016, CNN2014, LSTM_CNN2015, RGBD2021}.
However, most existing FAS datasets are relatively small and lack data diversity. No individual dataset can cover all types of presentation attacks and various environments, which leads to domain bias for the FAS training. As a result, models trained on a single FAS dataset usually exhibit poor generalization on unseen FAS datasets~\cite{SiW2018, MADDG_CVPR2019}.

To address this challenge, many efforts are devoted to domain generalization (DG) techniques to enhance the model's generalization to unseen domains. Generally, the model is trained on multiple FAS datasets simultaneously to seek a discriminative domain-invariant feature for liveness detection. Some works resort to aligning the FAS features from different domains to make them domain indistinguishable~\cite{MADDG_CVPR2019, DualReweighting2021, SSDG_CVPR2020, PatchNet_CVPR2022}. To further align the FAS feature, other interesting works focus on disentangling the domain-invariant and the domain-related features~\cite{disentangle_attention_ANRL2021, disentangle2022, SSAN_CVPR2022}. Besides, meta-learning is also utilized to help the model obtain the domain-invariant feature via simulating domain shift scenarios in the training stage~\cite{NAS_FAS2020, meta2020, fine_meta2020, SDA2021}. Although using domain generalization techniques relieves the lack of data diversity by training on multiple datasets, the capabilities of these methods are still constrained by the limited presentation attack types in the training datasets.
Actually, we find numerous real faces can be easily achieved under various conditions and assume that effectively using these real faces is a key to improving the generalization of the FAS model. However, we find it non-trivial to utilize these data, as the straightforward addition of the large-scale real faces to the training data can lead to a performance decline in existing DG-based methods. We believe this might be attributed to the gap in the distribution between the additional real faces and the test FAS dataset, which introduces domain bias to the detection model.

In this work, we propose an Anomalous cue Guided Face Anti-spoofing (AG-FAS) method that can effectively leverage additional real faces for improving cross-domain FAS. 
Specifically, we train a De-fake Face Generator (DFG) only on a large-scale real face dataset to gain the knowledge of what a real face should be like. The latent diffusion model based DFG can then remove the spoof trace of the input by adding noise during the forward process, i.e., the “de-fake” process, and restore a “real” version of the input via the generation process.
As we utilize the face identity of the input to guide the generation process, we can adopt larger steps during the forward process of DFG to thoroughly remove the fake patterns, leading to a more “real” output.
Consequently, we calculate the residual between the input and its “real” version, which is taken as the anomalous cue for the succeeding liveness detection. Then, we design an Off-real Attention Network (OA-Net) that utilizes the anomalous cue to allocate its attention toward the face region that deviates from the "real", i.e., the “off-real” region, achieving a more generalized FAS feature.
Moreover, as a Plug-and-Play feature extraction module, the proposed OA-Net can be seamlessly tailored into most existing DG-based FAS methods to further improve the model performance across different domains.
The experimental results show that the incorporation of this anomalous cue effectively enhances the performance of the model in unseen scenarios and unknown presentation attacks.

The main contributions of this work are summarized as follows:

\begin{itemize}
\item We propose an Anomalous cue Guided Face Anti-spoofing (AG-FAS) method, which can effectively leverage a large scale of additional real faces for improving model generalization on cross-domain FAS tasks.

\item We train a De-fake Face Generator (DFG) that obtains an effective anomalous cue for liveness detection. With the help of the anomalous cue, the proposed Off-real Attention Network (OA-Net) can focus on the off-real region of the input, leading to a more generalized FAS feature.

\item We provide theoretical analysis on the distinguishability of anomalous cues between real and fake inputs, which further guarantees the feasibility of the proposed framework.
\end{itemize}

\section{Related Work}\label{sec2}

Firstly, we present a brief overview of existing face anti-spoofing methods on the intra- and cross-dataset evaluations. Then, we introduce some anomaly detection works, which also utilize the model trained only on normal samples to detect the anomaly in the input. Finally, we review some typical works in diffusion models for unconditional and conditional image generation.

\subsection{Face Anti-spoofing Methods}
Traditional FAS methods focus on designing hand-craft descriptors that reflect different representations between real and spoof faces. The most commonly used hand-craft descriptors in the field of computer vision, such as LBP~\cite{LBP_HoG2012,LBP2015}, HoG~\cite{HoG2013}, SURF~\cite{Color2016}, SIFT~\cite{distortion2016}, etc, are all leveraged as the representations for detecting the face presentation attacks. Although the hand-craft descriptors always have good interpretability, they struggle when dealing with more complex attack scenarios. Later, deep neural networks are adopted to fix the FAS problem for their extraordinary feature learning capability. Commonly used neural networks like convolutional neural networks (CNNs) ~\cite{CNN2014,Depth_CNN_2017} and long short term memory (LSTM) ~\cite{LSTM_CNN2015} are all introduced to the field of FAS and achieve great progress. With the help of the deep neural networks, other modalities like depth maps~\cite{RGBD2021,Depth_CNN_2017} and reflection maps~\cite{reflection2019, reflection2020} can be used jointly as auxiliary information for face liveness detection. However, though the above FAS methods achieve excellent performance on intra-dataset evaluations, they cannot well handle the data in unseen domains due to the limitation of the training FAS dataset diversity. Some researchers attempt to employ domain adaptation methods to improve the model's performance in unseen domains.~\cite{Unsupervised_DA_Rose-Youtu2018, wang_improving_DA2019, SDA2021}. However, it remains uncertain which specific scenarios the model will encounter in real-world applications, and collecting data from all potential target datasets is challenging. 

To improve the performance of the FAS model in unseen domains, researchers have adopted domain generalization (DG) techniques that use multiple FAS datasets simultaneously to seek a discriminative domain-invariant feature for liveness detection. Adversarial learning is mostly used in DG-based works to extract a common FAS feature space for all the domains ~\cite{MADDG_CVPR2019, DualReweighting2021}. A later work points out that the feature diversity differentiates between the real and spoof faces and adopts an asymmetric triplet loss function~\cite{SSDG_CVPR2020}. Another typical work concatenates the domain-invariant and the domain-related features from different images and uses contrastive learning to disentangle these two features~\cite{SSAN_CVPR2022}. Recent work finds it difficult to directly construct a domain-invariant feature space and proposes to align the live-to-spoof transition~\cite{SA-FAS2023}.
Many other works further enhance cross-domain generalization in FAS tasks by focusing on various aspects such as gradient regulation~\cite{GAC-FAS_CVPR2024}, hierarchical relations in samples~\cite{hierarchical_CVPR2024}, and the utilization of testing data~\cite{Test-time-DG_CVPR2024}.
With the rise of large vision-language models such as CLIP~\cite{CLIP2021}, a new line of works emerges that leverage text information to guide models in performing face anti-spoofing~\cite{FLIP_ICCV2023,CFPL_CVPR2024}.
All existing FAS works based on domain generalization techniques achieve great improvement on several cross-dataset evaluation protocols. However, due to the lack of data diversity in existing FAS datasets, the capabilities of the previous FAS models are still constrained by the limited presentation attack types in the datasets. 

Some efforts are devoted to disentangling the attacks from the face by modeling the spoof traces~\cite{SpoofTrace_ECCV2020,SpoofTrace_TPAMI2022}, leading to robust liveness detection. Nevertheless, effectively modeling all types of spoof traces is a non-trivial task. In contrast, our method focuses on modeling the distribution of real faces using achievable large-scale data, thereby allowing for better generalization to unseen scenarios and unknown presentation attacks.

\subsection{Anomaly Detection Methods}
Anomaly detection is using the model trained solely on normal samples to identify whether a testing sample fits the same normal distribution in the training set or not. Early anomaly detection methods primarily focus on the feature space and determine the range of the normal features based on the local density. One commonly used approach is to train a one-class support vector machine (OC-SVM) to define the boundary between normal and anomalous samples in the feature space~\cite{one-class-SVM2016,one-class-SVM-OCT2016}. Later researchers use deep networks to fit the normal features into a minimum hypersphere or a Gaussian model~\cite{Deep_SVDD2018,Patch-SVDD2020,GMM_for_AD2018}.

Another mainstream anomaly detection method is based on image reconstruction. The model trained only on the normal samples is required to reconstruct the “normal” version image of the given input image. The difference between the generated “normal” image and the input image shows how close the input is to the normal distribution. A typical work based on autoencoder uses a memory bank strategy to prevent the model from constructing an identical image as the input~\cite{MemAE2019}. 
Subsequently, generative adversarial networks (GANs) are employed to perform the function of image reconstruction~\cite{AnoGAN2017,f-AnoGAN2019}. Since GAN training involves both an image encoder and decoder, GANs can similarly function as autoencoders to reconstruct the “normal” version of a given input image. Additionally, the adversarial learning in GANs allows for better reconstruction quality compared to most autoencoders, thereby improving anomaly detection performance. 
More recent anomaly detection works begin to explore the use of diffusion models for image reconstruction~\cite{AnoDDPM2022}. Unlike diffusion models used for unconditional image generation, here the model does not generate images directly from random noise. Instead, it first adds noise to the test image for a certain number of steps to remove the anomaly information. The model can then reconstruct a high-quality “normal” image based on the partially noised test image. This process, which resembles purifying the image, has also been used to defend against adversarial attacks in images~\cite{puri_diffusion_ICML2022}.

The above idea of reconstructing normal samples and obtaining anomalous cues is also applied to the field of liveness detection. A pioneering work~\cite{despoofing_IJCB2023} attempts to utilize a de-spoofing diffusion model (DDM) to restore real faces from the input. The differences between the restored image and the input are used for detecting the presentation attacks. However, this work still faces many limitations, such as the trade-off between removing the spoof information and preserving the original face, as well as the effective utilization of the anomalous cues. 
Compared to DDM, we propose to utilize the face identity of the input for getting more distinguishable anomalous cues and introduce an attention-based feature extractor to better leverage the hints from the off-real region, resulting in a more generalized FAS feature.

\subsection{Diffusion Model}
Diffusion model is a kind of multi-step generative model~\cite{DDPM2020, DDPM2021_classifier_guidance}, which gains widespread usage due to its ability to produce high-quality images. The early diffusion model is capable of generating from Gaussian noise unconditionally to obtain images that follow the distribution of the training dataset~\cite{DDPM2020}. During the training process, the clean images $x_0$ are first progressively added noise for $n$ steps towards Gaussian noise $x_n$. Subsequently, given a step number $t$ and the corresponding noisy image $x_t$ as input, the model is trained to reconstruct the noisy image $x_{t-1}$ at step $t-1$. 
In the practical implementation of the paper~\cite{DDPM2020}, the target model output is the noise $\epsilon$ added to the image. The training objective can be simplified to:
\begin{equation} \begin{aligned}
L_{DM} =  \mathbb{E}_{x,\epsilon \in \mathcal{N}(0,1),t}\Vert\epsilon-\epsilon_\theta(x_t,t)\Vert_2^2.
\end{aligned} \end{equation}

Subsequent researchers modify the structure of the diffusion model to accept conditional inputs $c$ and turn the diffusion model into a conditional generative model~\cite{classifier-free2022}. The training objective for the conditional diffusion model is as follows:
\begin{equation} \begin{aligned}
L_{CDM} =  \mathbb{E}_{x,c,\epsilon \in \mathcal{N}(0,1),t}\Vert\epsilon-\epsilon_\theta(x_t,t,c)\Vert_2^2.
\end{aligned} \end{equation}

Recently, it is discovered that training the diffusion model in the latent space yields higher efficiency and better generative results~\cite{LDM2022}. The training objective is as follows:
\begin{equation} \begin{aligned}
L_{LDM} =  \mathbb{E}_{E(x),y,\epsilon \in \mathcal{N}(0,1),t}\Vert\epsilon-\epsilon_\theta(z_t,t,\tau(y))\Vert_2^2,
\label{eq:ldm}
\end{aligned} \end{equation}
where $E$ is an encoder that embeds the input $x$ into the latent space feature $z_0$ and then obtains the noisy $z_t$ via the forward process. $\tau$ is another encoder which projects the conditional input $y$ to an intermediate representation.

\begin{figure*}[!ht]%
\centering
\includegraphics[width=0.8\linewidth]{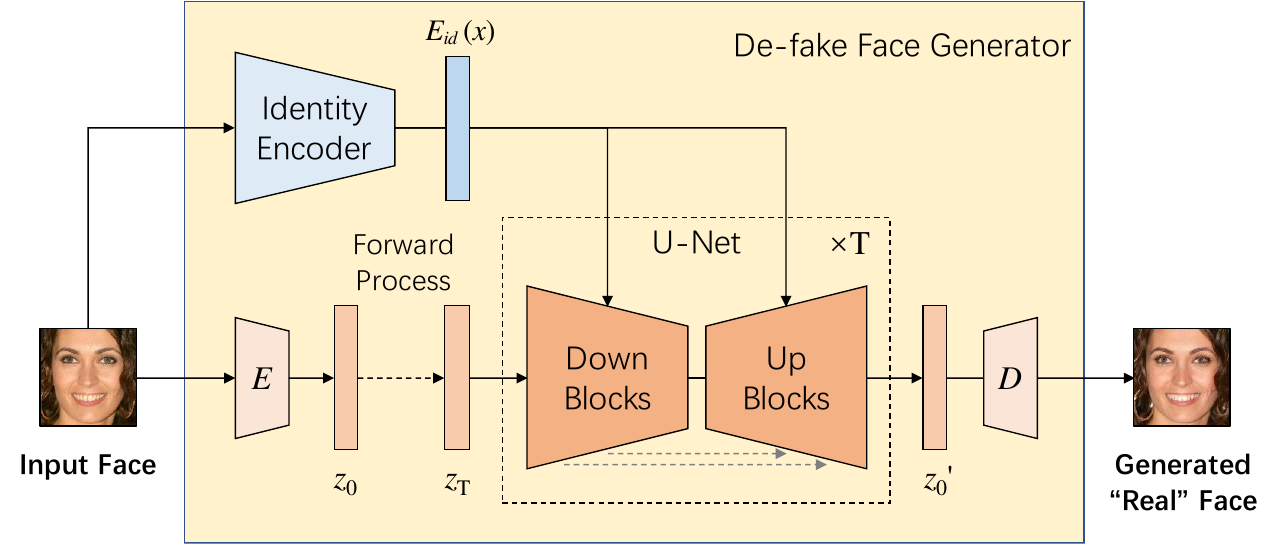}
\caption{Structure of the proposed De-fake Face Generator (DFG). The DFG trained on a large-scale real face dataset can generate a corresponding “real” face of any given input face when taking the input's identity feature $E_{id}(x)$ as guidance.}\label{fig:DFG}
\end{figure*}

\section{Method}
\subsection{Overview}
\begin{figure}[t]%
\centering
\includegraphics[width=0.4\textwidth]{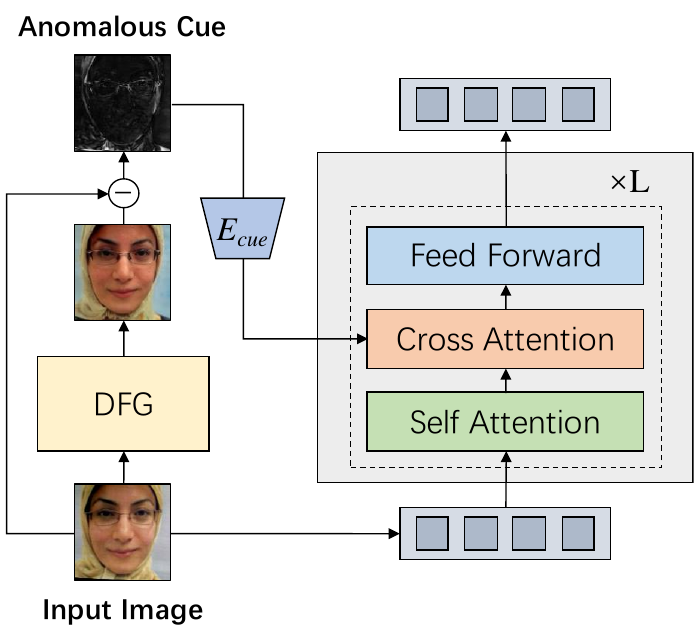}
\caption{Structure of the Off-real Attention Network (OA-Net). We take the residual of the input and the corresponding image generated by DFG as the anomalous cue, which guides the OA-Net to obtain a more robust FAS feature via the cross-attention module.}\label{fig:feature_extractor}
\end{figure}

Our method consists of two key modules, i.e., De-fake Face Generator (DFG) and Off-real Attention Network (OA-Net), as illustrated in \Cref{fig:DFG} and \Cref{fig:feature_extractor}, respectively.
We first train a De-fake Face Generator (DFG) on a large-scale real face dataset. The trained DFG is capable of generating a “real” version face of the input when taking the input face identity feature $E_{id}(x)$ as guidance. We obtain an anomalous cue of the input image by calculating the residual between the generated “real” face image and the input. The anomalous cue is then used as a hint to guide the Off-real Attention Network (OA-Net) to explore more robust FAS features. As a Plug-and-Play feature extraction module, the OA-Net can be applied in most existing domain generalization (DG) methods such as SSDG~\cite{SSDG_CVPR2020} to further enhance the generalization ability of the learned feature.

\subsection{De-fake Face Generator}
Firstly, we employ a dataset that contains only real faces to train a De-fake Face Generator (DFG), as shown in~\Cref{fig:DFG}, aiming to obtain a model that can generate corresponding “real” faces when given any faces as input. The real face dataset can be collected from the Internet or existing open-source facial datasets.

As for the backbone of our model, we choose Latent Diffusion Model (LDM)~\cite{LDM2022} for its excellent generation performance.
We propose an identity branch to obtain the conditional input for the reverse process in order to enhance the consistency between the input and the generated face. Therefore, compared to using the LDM alone, our DFG can capture better anomalous cues, effectively reducing noise from inaccurate face reconstruction caused by identity mismatches between the reconstructed face and the input face.
The training objective for DFG is as follows:
\begin{equation} \begin{aligned}
L_{DFG} =  \mathbb{E}_{x,E(x),\epsilon \in \mathcal{N}(0,1),t}\Vert\epsilon-\epsilon_\theta(z_t,t,E_{id}(x))\Vert_2^2,
\label{eq:L_DFG}
\end{aligned} \end{equation}
where $x$ is the input image, $E$ is the encoder which embeds the input $x$ into the clean latent code $z_0$, $z_t$ is the noisy latent code obtained via the forward process, $\epsilon$ is the Gaussian noise of the forward process, $\epsilon_\theta$ is the U-Net trained to predict the noise $\epsilon$, $t$ is the time step, and $E_{id}$ denotes the identity feature extractor. 
Specifically, an Arcface encoder~\cite{arcface2019} is used as the identity feature extractor to enhance the quality of the extracted features.
During training, the parameters of the encoder $E$ and the identity feature extractor $E_{id}$ are fixed, and only the parameters of the U-Net $\epsilon_\theta$ are optimized.

Once the U-Net is trained, the DFG can be used to generate “real” faces corresponding to the given faces. Specifically, the DFG first embeds the input $x$ into the clean latent code $z_0$ and obtains the noisy $z_{\hat{t}}$ via a determined number of diffusion steps $\hat{t}$. Then, the U-Net takes the identity feature $E_{id}(x)$ of the input face as the conditional input and progressively generates the latent code $z_0'$. Finally, we use the decoder $D$ in LDM to obtain the “real” face $x_0'$ from the latent code $z_0'$.

\subsection{Off-real Attention Network}
After obtaining a well-trained DFG, we generate a corresponding “real” face for every image in the FAS dataset. The residual between the generated “real” face and the input face is taken as the anomalous cue which shows how different the input is from the real face distribution.

We then design an Off-real Attention Network (OA-Net), as shown in \Cref{fig:feature_extractor}, to learn more robust FAS features with the help of the anomalous cue. 
The anomalous cue of the generated image is embedded by an anomalous cue encoder $E_{cue}$, utilizing a Resnet18~\cite{ResNet2016} backbone.
Meanwhile, the original input is preprocessed and put into the ViT-based model with $L$ layers. 

In order to exploit the anomalous cue, we follow the inspiration from the language model \cite{attention_is_all_you_need2017} and propose to add a cross-attention module in each layer of the model. 
When extracting the FAS feature, the OA-Net can obtain hints about the location of the off-real region by querying the embedded key and value of the anomalous cue at different layers. Compared to directly using ViT for feature extraction, focusing on off-real regions helps the OA-Net more effectively excavate robust FAS features.
We initialize the weight of the last linear layer of the cross-attention module to be zero, so the mainstem network does not receive information from the anomalous cue branch at the beginning. The model will gradually learn to exploit the hints from the anomalous cue branch during the training process.

\begin{figure}[t]%
\centering
\includegraphics[width=0.5\textwidth]{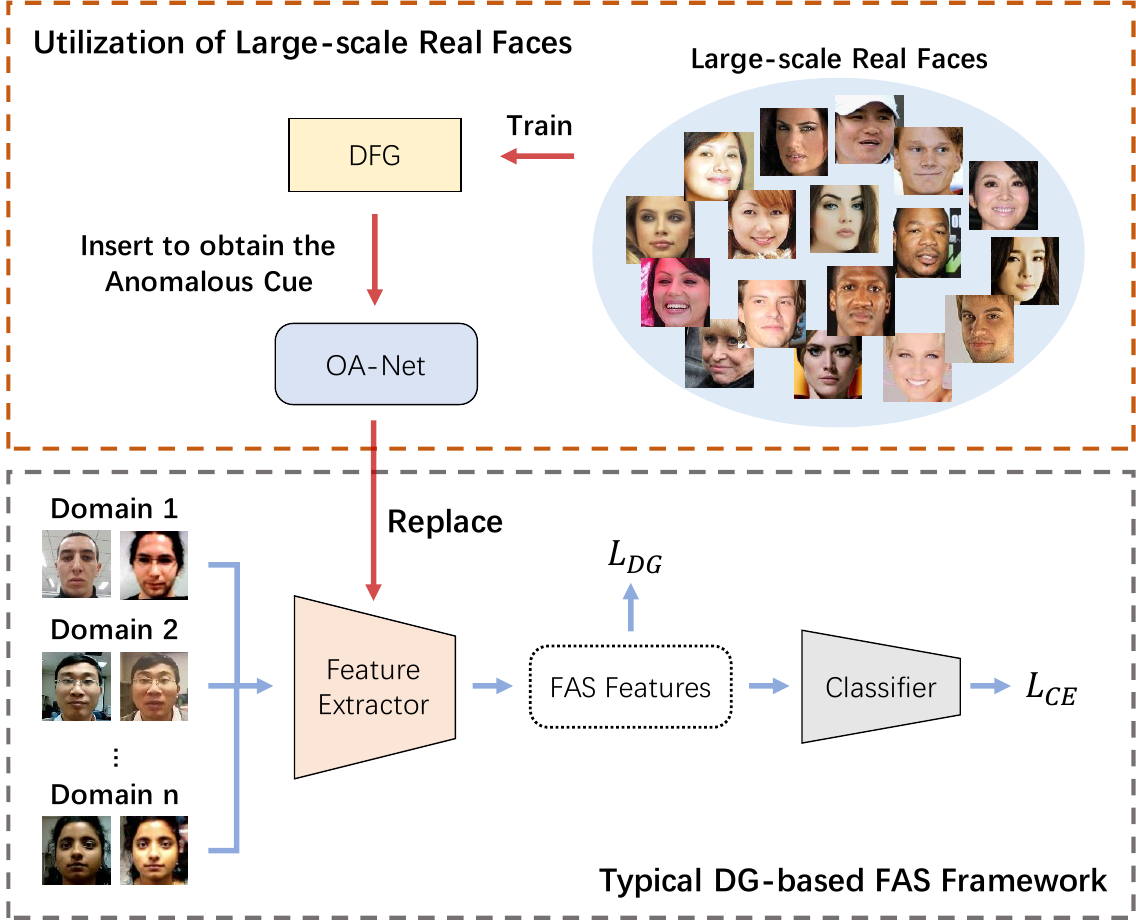}
\caption{The overall application diagram of AG-FAS, where OA-Net is integrated with existing DG-based FAS methods as a Plug-and-Play feature extraction module. $L_{DG}$ represents the feature constraint losses used in DG-based FAS methods, such as the adversarial loss and the asymmetric triplet loss in SSDG~\cite{SSDG_CVPR2020}.}\label{fig:overall}
\end{figure}

Finally, the extracted FAS feature is fed into a classifier to distinguish between real and fake faces. The classifier is a linear model implemented with a fully connected layer.
As a Plug-and-Play feature extraction module, the proposed OA-Net can be combined with most existing DG-based FAS methods,
as shown in \Cref{fig:overall}, further improving the generalization of the extracted FAS feature.

\section{Theoretical Analysis}
In this part, we provide theoretical guarantees for the distinguishability of the anomalous cues obtained from real and fake faces, which demonstrates the feasibility of the proposed framework.  
Considering the latent space learned by the autoencoding model in LDM~\cite{LDM2022} is perceptually equivalent to the image space, we also conduct our theoretical analysis in the latent space for simplification.

The theoretical analysis is performed in two directions: first, we prove the invariance of the latent codes for real faces before and after processing through the DFG, which indicates the corresponding anomalous cues tend to zero; then, we demonstrate that the output latent codes will differ from the fake input codes that contain spoof traces, resulting in noticeable anomalous cues for fake faces.

\subsection{Anomalous Cue of Real Face}

First of all, the following theorem reveals the invariance of the latent codes for real faces.

\textbf{Theorem 1} \textit{For a latent code $z_0$ of a real face conforming to the distribution of the DFG training set and a given reconstruction error $\Delta z_0 > 0$, there exists a threshold time steps $t_{thr}$, such that when $t<t_{thr}$:
\begin{equation} \begin{aligned}
\mathbb{E}_{z_t \in q(z_t|z_0)}\Vert z_0-z_0' \Vert_2^2 < \Delta z_0,
\label{eq:theorem_1}
\end{aligned} \end{equation}
where $q(z_t|z_0)$ is the forward process, $z_0'$ is the final output of the DFG reverse process using the sampled noisy latent code $z_t$.}

Theorem 1 states that, by limiting the time steps of the forward process, the DFG model can effectively restore the original latent code $z_0$ based on the noisy latent code $z_t$ obtained from the forward process, which indicates the anomalous cue of a real input tends to be a matrix approaching $\mathbf{0}$.

Before introducing the proof for Theorem 1, we present the following assumption to ensure that the DFG model is sufficiently trained:

\textbf{Assumption 1} \textit{We assume that the model is sufficiently trained. For a latent code $z_0$ of a real face conforming to the distribution of the DFG training set and a given predictive error $\Delta \epsilon > 0$, there exists a threshold time step $t_{thr}$, such that when $t_r \leq t<t_{thr}$:
\begin{equation} \begin{aligned}
\mathbb{E}_{z_t \in q(z_t|z_0)}\Vert\epsilon-\epsilon_\theta(z_{t_r}',t_r,E_{id}(x))\Vert_2^2 < \Delta \epsilon,
\end{aligned} \end{equation}
where $q(z_t|z_0)$ is the forward process, $z_t$ is the noisy latent code, $\epsilon$ is the Gaussian noise of the forward process, $z_{t_r}'$ is the generated result in the reverse process at time step $t_r$, and $E_{id}(x)$ is the identity feature of the corresponding input. }

The assumption implies that, within a certain time step, the well-trained model can effectively predict the added Gaussian noise $\epsilon$ in the forward process. With this assumption, we give the following proof of Theorem 1. 

\textbf{Proof of Theorem 1}: Follow the definition in \cite{DDIM2021}, the reverse process to generate $z_{t-1}'$ from $z_t'$ in the latent diffusion model is as follows:
\begin{equation} \begin{aligned}
z_{t-1}'&=\sqrt{\alpha_{t-1}}\left( \frac{z_t'-\sqrt{1-\alpha_t} \cdot \epsilon_\theta^t}{\sqrt{\alpha_t}}\right) \\
&+\sqrt{1-\alpha_{t-1}-\sigma_t^2} \cdot \epsilon_\theta^t + \sigma_t \epsilon_t,
\end{aligned} \end{equation}
where $\epsilon_\theta^t$ represents $\epsilon_\theta(z_t',t,E_{id}(x))$ for simplification, $\alpha_0$ is defined as 1, $\epsilon_t$ is a standard Gaussian noise, and $\sigma_t$ is a hyperparameter that is set as 0 in our DFG to obtain a denoising diffusion implicit model (DDIM)~\cite{DDIM2021}.

Thus, we can simplify the reverse process and express the generation output $z_0'$ in a form represented by the noisy latent code $z_t$ as follows:
\begin{equation} \begin{aligned}
z_0'&=\sqrt{\alpha_{0}}\left( \frac{z_1'-\sqrt{1-\alpha_1} \cdot \epsilon_\theta^1}{\sqrt{\alpha_1}}\right) +\sqrt{1-\alpha_0} \cdot \epsilon_\theta^1 \\
&= \frac{\sqrt{\alpha_0}}{\sqrt{\alpha_1}} z_1' - \left( \frac{\sqrt{\alpha_0}}{\sqrt{\alpha_1}}\sqrt{1-\alpha_1} - \sqrt{1-\alpha_0} \right) \cdot \epsilon_\theta^1 \\
&= \frac{\sqrt{\alpha_0}}{\sqrt{\alpha_2}} z_2' - \left( \frac{\sqrt{\alpha_0}}{\sqrt{\alpha_2}}\sqrt{1-\alpha_2} - \frac{\sqrt{\alpha_0}}{\sqrt{\alpha_1}}\sqrt{1-\alpha_1} \right) \cdot \epsilon_\theta^2 \\
&-  \left( \frac{\sqrt{\alpha_0}}{\sqrt{\alpha_1}}\sqrt{1-\alpha_1} - \sqrt{1-\alpha_0} \right) \cdot \epsilon_\theta^1 \\
&= ...... \\
&= \frac{\sqrt{\alpha_0}}{\sqrt{\alpha_t}} z_t' - \sum_{i=1}^t \left( \frac{\sqrt{\alpha_0}}{\sqrt{\alpha_i}}\sqrt{1-\alpha_i} \right. \\
&\left. - \frac{\sqrt{\alpha_0}}{\sqrt{\alpha_{i-1}}}\sqrt{1-\alpha_{i-1}} \right) \cdot \epsilon_\theta^i.
\end{aligned} \end{equation}

Since $\alpha_0$ is defined as 1 and $z_t'$ is actually the noisy latent code $z_t$ obtained from the forward process, the generation output $z_0'$ can be expressed as:
\begin{equation} \begin{aligned}
z_0' = \frac{1}{\sqrt{\alpha_t}} z_t - \sum_{i=1}^t \left( \frac{\sqrt{1-\alpha_i}}{\sqrt{\alpha_i}} - \frac{\sqrt{1-\alpha_{i-1}}}{\sqrt{\alpha_{i-1}}} \right) \cdot \epsilon_\theta^i.
\label{eq:z_0'}
\end{aligned} \end{equation}

With the definition of the forward process $q(z_t|z_0)$, the noisy latent code $z_t$ is obtained as:
\begin{equation} \begin{aligned}
z_t = \sqrt{\alpha_t}z_0 + \sqrt{1-\alpha_t}\epsilon,
\end{aligned} \end{equation}
where $\epsilon$ is the standard Gaussian noise used in the forward process. The expression of $z_0$ can then be represented by $z_t$ as follows:
\begin{equation} \begin{aligned}
z_0 =  \frac{1}{\sqrt{\alpha_t}}z_t - \frac{\sqrt{1-\alpha_t}}{\sqrt{\alpha_t}} \epsilon.
\label{eq:z_0}
\end{aligned} \end{equation}

Taking the results from \Cref{eq:z_0'} and \Cref{eq:z_0}, we can calculate the reconstruction error for a given latent code $z_0$ of a real face as follows:
\begin{equation} \begin{aligned}
&\mathbb{E}_{z_t \in q(z_t|z_0)}\Vert z_0-z_0' \Vert_2^2 \\
&=\mathbb{E}_{z_t}\Vert \frac{1}{\sqrt{\alpha_t}}z_t - \frac{\sqrt{1-\alpha_t}}{\sqrt{\alpha_t}} \epsilon - \frac{1}{\sqrt{\alpha_t}} z_t \\
&+ \sum_{i=1}^t \left( \frac{\sqrt{1-\alpha_i}}{\sqrt{\alpha_i}} - \frac{\sqrt{1-\alpha_{i-1}}}{\sqrt{\alpha_{i-1}}} \right) \cdot \epsilon_\theta^i \Vert_2^2 \\
&=\mathbb{E}_{z_t}\Vert \sum_{i=1}^t \left( \frac{\sqrt{1-\alpha_i}}{\sqrt{\alpha_i}} - \frac{\sqrt{1-\alpha_{i-1}}}{\sqrt{\alpha_{i-1}}} \right) \cdot \left( \epsilon_\theta^i - \epsilon \right) \Vert_2^2 \\
&=\sum_{i=1}^t \left[ \left( \frac{\sqrt{1-\alpha_i}}{\sqrt{\alpha_i}} - \frac{\sqrt{1-\alpha_{i-1}}}{\sqrt{\alpha_{i-1}}} \right) \mathbb{E}_{z_t}\Vert \epsilon_\theta^i - \epsilon \Vert_2^2 \right].
\end{aligned} \end{equation}

After applying Assumption 1, the upper bound of the reconstruction error turns out to be:
\begin{equation} \begin{aligned}
\mathbb{E}_{z_t}\Vert z_0-z_0' \Vert_2^2 &< \sum_{i=1}^t \left( \frac{\sqrt{1-\alpha_i}}{\sqrt{\alpha_i}} - \frac{\sqrt{1-\alpha_{i-1}}}{\sqrt{\alpha_{i-1}}} \right) \Delta \epsilon \\
&= \frac{\sqrt{1-\alpha_t}}{\sqrt{\alpha_t}} \Delta \epsilon.
\label{eq:upper_bound}
\end{aligned} \end{equation}

Compared the upper bound with \Cref{eq:theorem_1}, we can conclude that Theorem 1 holds with a proper $t_{thr}$ that satisfies both the Assumption 1 and the following equation:
\begin{equation} \begin{aligned}
\Delta \epsilon \leq \frac{\sqrt{\alpha_{t_{thr}}}}{\sqrt{1-\alpha_{t_{thr}}}} \Delta z_0 < \frac{\sqrt{\alpha_t}}{\sqrt{1-\alpha_t}} \Delta z_0.
\end{aligned} \end{equation}

It is worth noting that directly substituting the noise $\epsilon$ in \Cref{eq:z_0} with the predicted result $\epsilon_\theta(z_t,t,E_{id}(x))$ can also yield the upper bound in \Cref{eq:upper_bound}, which implies that a single-step reverse process can also obtain an upper bound on the reconstruction error.

\subsection{Analysis for Forward Process}

Before we turn to analyze the anomalous cues of fake faces, we need the following preliminary theorem.

\textbf{Theorem 2} \textit{For the forward process of a latent diffusion model $q(z_t|z_0)$, if the length $T$ of the forward process is sufficiently large, for a given $\epsilon_{KL}>0$ and two latent codes $z_0^{r}, z_0^{f}$ of the real and fake faces from the same identity, there exists a time step $t$ such that:
\begin{equation} \begin{aligned}
D_{KL}\{q(z_t^{r}|z_0^{r}), q(z_t^{f}|z_0^{f})\} < \epsilon_{KL},
\label{eq:theorem_2}
\end{aligned} \end{equation}
That is, the KL divergence between the latent distributions of the real and fake faces is sufficiently small under a certain time step $t$, making these two distributions indistinguishable.}

The theorem indicates that the fake patterns in the latent codes can be removed through the forward process, which makes them indistinguishable from the latent codes of the real faces. 

\textbf{Proof of Theorem 2}: Follow the definition in \cite{DDIM2021}, the forward process of a latent diffusion model is as follows:
\begin{equation} \begin{aligned}
q(z_t|z_0):=\mathcal{N}(z_t;\sqrt{\alpha_t}z_0,(1-\alpha_t)I),
\label{eq:app_forward_process}
\end{aligned} \end{equation}
where $\alpha_1,...,\alpha_T$ represent a series of monotonically decreasing coefficients, ensuring that $0<\alpha_T<...<\alpha_1<1$. 

The KL divergence between the latent distributions of the real and fake faces can then be computed:
\begin{equation} \begin{aligned}
&D_{KL}\{q(z_t^{r}|z_0^{r}), q(z_t^{f}|z_0^{f})\} \\
&= D_{KL}\{\mathcal{N}(z_t^r;\sqrt{\alpha_t}z_0^r,(1-\alpha_t)I), \\
&\quad\quad\quad\quad \mathcal{N}(z_t^f;\sqrt{\alpha_t}z_0^f,(1-\alpha_t)I)\} \\
&=\log\frac{\sqrt{1-\alpha_t}}{\sqrt{1-\alpha_t}} + \frac{(1-\alpha_t)+(\sqrt{\alpha_t}(z_0^r-z_0^f))^2}{2(1-\alpha_t)} - \frac{1}{2} \\
&=\frac{\alpha_t}{2(1-\alpha_t)}(z_0^r-z_0^f)^2.
\end{aligned} \end{equation}

It can be observed that the KL divergence monotonically decreases as $\alpha_t$ decreases, and as $\alpha_t$ approaches 0, the KL divergence also approaches 0. 

Taking the result into \Cref{eq:theorem_2}, we obtain the following condition required by Theorem 2:
\begin{equation} \begin{aligned}
\frac{\alpha_t}{2(1-\alpha_t)}(z_0^r-z_0^f)^2 < \epsilon_{KL}.
\end{aligned} \end{equation}

After simplification, the range of the available value for $\alpha_t$ is as follows:
\begin{equation} \begin{aligned}
\alpha_t < \frac{2\epsilon_{KL}}{2\epsilon_{KL}+(z_0^r-z_0^f)^2}.
\end{aligned} \end{equation}

Since the length $T$ is sufficiently large, $\alpha_T$ is set close to 0, which implies there exists a time step $t$ and its corresponding $\alpha_t$ such that:
\begin{equation} \begin{aligned}
\alpha_T \leq \alpha_t < \frac{2\epsilon_{KL}}{2\epsilon_{KL}+(z_0^r-z_0^f)^2},
\end{aligned} \end{equation}
which means Theorem 2 holds.

The formula also shows that the smaller the difference between the two latent codes $z_0^{r}$ and $z_0^{f}$, the larger the upper limit for $\alpha_t$, and consequently, the smaller the time step $t$ is required in the forward process. When the two latent codes are equal, $\alpha_1 < 1$ with the time step $t=1$ just satisfies the condition. 

\subsection{Anomalous Cue of Fake Face}

Theorem 2 shows that the forward process can remove the fake patterns, and Theorem 1 shows that the model can reconstruct the clean latent code of a real face from the noisy latent code. Combining Theorem 1 and Theorem 2, when the time step $t$ satisfies all the premises, we can obtain the following derivation for a fake input. 

We first employ Theorem 2 and replace the noisy fake latent code $z_t^f$ with the noisy real latent code $z_t^r$:
\begin{equation} \begin{aligned}
&\mathbb{E}_{z_t^f \in q(z_t^{f}|z_0^{f})}\Vert z_0^f-z_0' \Vert_2^2\approx \mathbb{E}_{z_t^r \in q(z_t^{r}|z_0^{r})}\Vert z_0^f-z_0' \Vert_2^2,
\label{eq:fake_cue_1}
\end{aligned} \end{equation}
where $z_0'$ is the output of the reverse process determined by the sampled noisy latent code $z_t^f$ or $z_t^r$.

Then we utilize the triangle inequality and employ Theorem 1 to obtain the lower bound of the reconstruction error:
\begin{equation} \begin{aligned}
&\mathbb{E}_{z_t^r \in q(z_t^{r}|z_0^{r})}\Vert z_0^f-z_0' \Vert_2^2 \\
&\geq \mathbb{E}_{z_t^r \in q(z_t^{r}|z_0^{r})} \left[ \Vert z_0^f-z_0^r \Vert_2^2-\Vert z_0^r-z_0' \Vert_2^2 \right] \\
&= \Vert z_0^f-z_0^r \Vert_2^2 - \mathbb{E}_{z_t^r \in q(z_t^{r}|z_0^{r})}\Vert z_0^r-z_0' \Vert_2^2 \\
&> \Vert z_0^f-z_0^r \Vert_2^2 - \Delta z_0,
\label{eq:fake_cue_2}
\end{aligned} \end{equation}

The derivation above indicates that when the input is a fake image, the residual between the input $z_0^f$ and the output $z_0'$ of the DFG will reflect the input's spoof traces $\Vert z_0^f-z_0^r \Vert_2^2$, resulting in a noticeable anomalous cue. Combining the result from Theorem 1, we prove the distinguishability between the anomalous cues obtained from real and fake faces.

\subsection{Discussion}
We find there exists a trade-off on the time step $t$ between Theorem 1 and Theorem 2: Theorem 1 requires a small time step to ensure the reconstruction of the latent code while Theorem 2 requires a large time step to remove the fake patterns. 

The trade-off indicates that expanding the available range of the time step $t$ is the key to ensuring the practical performance of the DFG.
Since the required $t$ in Theorem 2 is all determined by the difference between the input and its corresponding real face, we turn to study whether the threshold time step $t_{thr}$ in Theorem 1 can take a larger value.
The proof of Theorem 1 indicates that, while keeping the reconstruction error of the latent code $\Delta z_0$ constant, increasing the value of the threshold time step $t_{thr}$ requires decreasing the prediction error of the noise $\Delta \epsilon$.
Inspired by this observation, we utilize the identity feature as the guidance for the DFG model to obtain lower predictive error $\Delta \epsilon$, which allows us to choose a larger time step $\hat{t}$ for the DFG. Note that introducing the identity feature does not affect the derivation in \Cref{eq:fake_cue_1} because the identity features are consistent within real and fake faces of the same people.

The ablation study of the trade-off on the time step $\hat{t}$ for the DFG is conducted in \Cref{sec:compare_diffusion_steps} and the results are shown in \Cref{tab:compare_diffusion_steps}.

\section{Experiments}

\subsection{Dataset}

We adopt four typical datasets to construct the most commonly used cross-domain benchmark: MSU-MFSD~\cite{MSU2015}(denoted as M), CASIA-FASD~\cite{CASIA2012}(denoted as C), Idiap Replay-Attack~\cite{REPLAY2012}(denoted as I), and OULU-NPU~\cite{OULU2017}(denoted as O). Additionally, we introduce five other datasets that are commonly used in FAS evaluations: HKBU-MARs~\cite{HKBU2016}, WFFD~\cite{WFFD2020}, Rose-Youtu~\cite{Unsupervised_DA_Rose-Youtu2018}, SiW~\cite{SiW2018}, WMCA~\cite{WMCA2019}. Based on all nine datasets, we conduct an experiment to further evaluate the generalization of our method with unseen scenarios and unknown presentation attacks. Some of the datasets above are in video form, and we uniformly sample 10 frames from each video in the time dimension for training and testing. 

For the training of the De-fake Face Generator (DFG), we use the data from FFHQ~\cite{FFHQ2019} and the real faces from CelebA-Spoof~\cite{CelebA-Spoof2020}. After filtering out images with deviated facial poses, we collect about 130,000 real faces in total.

We apply the face alignment to all the data using MTCNN~\cite{MTCNN2016}. After alignment, all the images are cropped and resized to a shape of $224\times 224$.

\begin{table*}[t]
\centering
\caption{Evaluations on the Leave-One-Out Protocol}\label{tab:LOO}
\resizebox{\textwidth}{!}{
\begin{tabular}{c|c|c|c|c|c|c|c|c|c|c}
\hline
                & \multicolumn{2}{c|}{\textbf{O\&C\&I   to M}} & \multicolumn{2}{c|}{\textbf{O\&M\&I   to C}} & \multicolumn{2}{c|}{\textbf{O\&C\&M   to I}} & \multicolumn{2}{c|}{\textbf{I\&C\&M   to O}} & \multicolumn{2}{c}{\textbf{Mean}} \\
\cline{2-11}
                & HTER                 & AUC                  & HTER                 & AUC                  & HTER                 & AUC                  & HTER                 & AUC                  & HTER            & AUC             \\
\hline
MADDG~\cite{MADDG_CVPR2019}           & 17.69                & 88.06                & 24.50                & 84.51                & 22.19                & 84.99                & 27.98                & 80.02                & 23.09           & 84.40           \\
SSDG-M~\cite{SSDG_CVPR2020}          & 16.67                & 90.47                & 23.11                & 85.45                & 18.21                & 94.61                & 25.17                & 81.83                & 20.79           & 88.09           \\
NAS-FAS~\cite{NAS_FAS2020}         & 19.53                & 88.63                & 16.54                & 90.18                & 14.51                & 93.84                & 13.80                & 93.43                & 16.10           & 91.52           \\
D2AM~\cite{MixtureDomain_DDAM_AAAI2021}            & 12.70                & 95.66                & 20.98                & 85.58                & 15.43                & 91.22                & 15.27                & 90.87                & 16.10           & 90.83           \\
SDA~\cite{SDA2021}             & 15.40                & 91.80                & 24.50                & 84.40                & 15.60                & 90.10                & 23.10                & 84.30                & 19.65           & 87.65           \\
ANRL~\cite{disentangle_attention_ANRL2021}            & 10.83                & 96.75                & 17.83                & 89.26                & 16.03                & 91.04                & 15.67                & 91.90                & 15.09           & 92.24           \\
SSDG-R~\cite{SSDG_CVPR2020}          & 7.38                 & 97.17                & 10.44                & 95.94                & 11.71                & 96.59                & 15.61                & 91.54                & 11.29           & 95.31           \\
SSAN-R~\cite{SSAN_CVPR2022}          & 6.67                 & \textbf{98.75}       & 10.00                & 96.67          & 8.88                 & 96.79                & 13.72                & 93.63                & 9.82            & 96.46           \\
PatchNet~\cite{PatchNet_CVPR2022}        & 7.10                 & \uline{98.46}          & 11.33                & 94.58                & 13.40                & 95.67                & 11.82                & 95.07                & 10.91           & 95.95           \\
SA-FAS~\cite{SA-FAS2023}          & \uline{5.95}           & 96.55                & 8.78           & 95.37                & \textbf{6.58}        & 97.54          & \uline{10.00}          & \uline{96.23}          & \uline{7.83}      & 96.42           \\
DDM~\cite{despoofing_IJCB2023}          & 14.29              & 91.80             & 37.78           & 68.22           & 20.14              & 81.88              & 29.31        & 78.19     & 25.38         & 80.02     \\
DDM*~\cite{despoofing_IJCB2023}          & \uline{5.95}               & 97.64             & \uline{6.67}            & \uline{98.15}           & 8.64               & \uline{97.75}              & 14.72        & 93.12     & 9.00          & \uline{96.66}     \\
\textbf{AG-FAS (Ours)} & \textbf{5.71}        & 98.03                & \textbf{5.44}        & \textbf{98.55}       & \uline{6.71}           & \textbf{98.23}       & \textbf{9.43}        & \textbf{96.62}       & \textbf{6.82}   & \textbf{97.86} \\
\hline
\end{tabular}%
}
\end{table*}

\subsection{Implementation Details}
For the training of the DFG, we utilize the standard Latent Diffusion Model~\cite{LDM2022} as the backbone. Arcface encoder~\cite{arcface2019} is used to get the face identity feature as the conditional guidance. The shape of the Arcface output before the pooling layer is $512\times7\times7$, which is reshaped to the size of $49\times512$ and then padded as $49\times768$ to fit the original conditional input of the U-Net in Latent Diffusion Model. During training, we set the batch size to 32 and train for a total of 150,000 steps on the real face dataset. The Adam optimizer is employed with a learning rate of 1e-05, a weight decay of 0.01, $\beta_1$ of 0.9, and $\beta_2$ of 0.999. The maximum number of the diffusion steps is set to 1000. The parameters of the Arcface model and the encoder in Latent Diffusion Model are kept fixed during training.

For the training of the FAS model, we take the ViT-base~\cite{vit2020} model as the backbone of the Off-real Attention Network (OA-Net). We then add the cross-attention modules used for introducing the anomalous cue after each self-attention module of the original ViT model. The input images are divided with a patch size of $16\times16$ and embedded into image tokens in the shape of $196\times768$. The anomalous cue of each image calculated by the DFG is first embedded by a cue encoder to the shape of $49\times512$ and then put in the cross-attention modules of the OA-Net to provide hints about the anomaly. The cue encoder adopts the structure of a Resnet18~\cite{ResNet2016}. On the whole, we train our model using SSDG~\cite{SSDG_CVPR2020} strategy and use the open-source code to ensure consistency and comparability.

\subsection{Leave-One-Out Experiments}

Following the Leave-One-Out protocol with the same settings in~\cite{SSDG_CVPR2020}, we each time use three of the four datasets (MSU-MFSD, CASIA-FASD, Idiap Replay-Attack, and OULU-NPU) for training, while the remaining one dataset is used for testing. We compare our method with the state-of-the-art (SOTA) methods and the experimental results are summarized in~\Cref{tab:LOO}.
It can be observed that our method outperforms all the existing SOTA methods. We attribute this to the utilization of the anomalous cue that guides the learning of the FAS feature. 

Note that DDM also employs the idea of reconstructing real faces, using residuals for liveness detection. However, because DDM is not designed for the domain generalization task, its performance is relatively poor. Therefore, we first introduce the SSDG strategy into DDM so it can better perform the DG tasks. Then, to ensure a fair comparison, we replace the backbone of DDM with the same one as ours.
The improved DDM method is denoted as “DDM*” in the table. 

The improved DDM exhibits good performance on this DG protocol, but it is still inferior to our method.
This is because we utilize the face identity as the guidance for the DFG model, which allows us to add noise for more steps in the forward process according to the discussion in our theoretical analysis. As a result, a larger time step leads to better elimination of spoof information in the input, thus making anomalous cues more distinguishable. Besides, compared to DDM which learns the feature from two separate branches for the input image and the residual, our OA-Net enables the network to better focus on the off-real region according to the anomalous cue via the cross-attention modules, thereby obtaining features more relevant to the fake patterns.

\begin{table}[t]
\centering
\caption{Results with limited source domains}\label{tab:limited_source_domains}
\resizebox{0.5\textwidth}{!}{
\begin{tabular}{c|c|c|c|c}
\hline
                & \multicolumn{2}{c|}{\textbf{M\&I to C}} & \multicolumn{2}{c}{\textbf{M\&I to O}}  \\
\cline{2-5}
                & HTER            & AUC              & HTER            & AUC              \\
\hline
MADDG~\cite{MADDG_CVPR2019}           & 41.02           & 64.33            & 39.35           & 65.10               \\
SSDG-M~\cite{SSDG_CVPR2020}          & 31.89           & 71.29            & 36.01           & 66.88             \\
D2AM~\cite{MixtureDomain_DDAM_AAAI2021}            & 32.65           & 72.04            & 27.70           & 75.36           \\
SSAN-M~\cite{SSAN_CVPR2022}          & 30.00           & 76.20            & 29.44           & 76.62          \\
SSDG-R~\cite{SSDG_CVPR2020}          & 20.56       & 85.42       & 21.39       & 87.43          \\
SSAN-R~\cite{SSAN_CVPR2022}          & 27.33           & 79.52            & 28.75           & 77.78           \\
SA-FAS~\cite{SA-FAS2023}          & 30.11           & 73.87            & 28.25           & 80.02          \\
\textbf{AG-FAS (Ours)} & \textbf{7.33}        & \textbf{96.99}       & \textbf{15.97}       & \textbf{91.10}  \\
\hline
\end{tabular}
}
\end{table}

\subsection{Experiments with Limited Source Domains}

We conducted cross-domain FAS experiments in a more challenging setting with limited-source domains. We use MSU-MFSD and Idiap Replay-Attack datasets for training and the rest two datasets for testing. The results are shown in~\Cref{tab:limited_source_domains}. Despite the limited availability of training data, our method outperforms all existing SOTA methods. This outcome demonstrates that the anomalous cue can still guide the FAS model to learn a more generalized feature even with limited source training data.

\subsection{Study DG-based Methods with Additional Data}
\begin{table*}[t]
\centering
\caption{Comparison with existing DG-based methods that use additional real faces (*: using additional real faces during training)}\label{tab:additional_data}
\resizebox{\textwidth}{!}{
\begin{tabular}{c|c|c|c|c|c|c|c|c|c|c}
\hline
                & \multicolumn{2}{c|}{\textbf{O\&C\&I   to M}} & \multicolumn{2}{c|}{\textbf{O\&M\&I   to C}} & \multicolumn{2}{c|}{\textbf{O\&C\&M   to I}} & \multicolumn{2}{c|}{\textbf{I\&C\&M   to O}} & \multicolumn{2}{c}{\textbf{Mean}} \\
\cline{2-11}
                & HTER                 & AUC                  & HTER                 & AUC                  & HTER                 & AUC                  & HTER                 & AUC                  & HTER            & AUC             \\
\hline
SSDG-R~\cite{SSDG_CVPR2020}          & 7.38                 & 97.17                & 10.44                & 95.94                & 11.71                & 96.59                & 15.61                & 91.54                & 11.29           & 95.31           \\
SSAN-R~\cite{SSAN_CVPR2022}          & 6.67                 & \textbf{98.75}       & 10.00       & 96.67                & 8.88                 & 96.79                & 13.72                & 93.63                & 9.82            & 96.46           \\
SA-FAS~\cite{SA-FAS2023}          & 5.95                 & 96.55                & 8.78                 & 95.37                & 6.58                 & 97.54                & 10.00                & 96.23                & 7.83            & 96.42           \\
\hline
SSDG-R*~\cite{SSDG_CVPR2020}          & 10.00                & 95.26                & 17.33                & 89.07                & \textbf{4.93}        & \textbf{99.28}       & 11.96                & 94.94                & 11.06         & 94.64          \\
SSAN-R*~\cite{SSAN_CVPR2022}          & 11.43                & 94.80                & 21.33                & 87.54                & 12.86                & 94.08                & 19.65                & 89.65                & 16.32         & 91.52          \\
SA-FAS*~\cite{SA-FAS2023}          & 25.48                & 82.05                & 24.44                & 84.42                & 22.86                & 84.76                & 25.04                & 82.96                & 24.45         & 83.55          \\
\textbf{AG-FAS*(Ours)} & \textbf{5.71}       & 98.03               & \textbf{5.44}       & \textbf{98.55}       & 6.714                & 98.23               & \textbf{9.43}       & \textbf{96.62}      & \textbf{6.82} & \textbf{97.86} \\
\hline
\end{tabular}%
}
\end{table*}
Since the proposed AG-FAS leverages additional real faces to help improve the FAS performance, we want to investigate whether existing DG-based methods can also benefit from these data. Specifically, we include the 130,000 additional real faces in the training dataset of the DG-based methods. 
To ensure balanced data sampling, in addition to maintaining an overall 1:1 ratio between real and fake faces during sampling, we also resample the real faces within each minibatch. This ensures a 1:1 ratio between the real faces from the original FAS dataset and the additional real faces.
The results of the experiments are summarized in~\Cref{tab:additional_data}. 
It can be observed that the performance of the DG-based methods does not show a significant improvement; instead, there is even a noticeable degradation in some methods.

This might be attributed to the gap in the FAS feature distribution between the additional real faces and the test FAS dataset. It is difficult to improve the model generalization by simply introducing additional real faces to existing DG-based methods. On the contrary, our AG-FAS learns an anomalous cue from the real faces, which is a more general pattern across different scenarios and thus helps improve the cross-dataset performance of the model, demonstrating the superiority of tackling the model generalization problem.

\begin{table*}[t]
\centering
\caption{Ablation study on different anomalous cue obtaining approaches (\dag: mask the background of the anomalous cue)}\label{tab:compare_anomaly_detection}
\resizebox{\textwidth}{!}{
\begin{tabular}{c|c|c|c|c|c|c|c|c|c|c}
\hline
                & \multicolumn{2}{c|}{\textbf{O\&C\&I   to M}} & \multicolumn{2}{c|}{\textbf{O\&M\&I   to C}} & \multicolumn{2}{c|}{\textbf{O\&C\&M   to I}} & \multicolumn{2}{c|}{\textbf{I\&C\&M   to O}} & \multicolumn{2}{c}{\textbf{Mean}} \\
\cline{2-11}
                & HTER                 & AUC                  & HTER                 & AUC                  & HTER                 & AUC                  & HTER                 & AUC                  & HTER            & AUC             \\
\hline
baseline      & 6.91                 & 97.06                & 8.00                 & 97.37                & 7.71                 & 97.96                & 16.11                & 92.43                & 9.68            & 96.20           \\
MemAE~\cite{MemAE2019}        & \textbf{4.29}        & \textbf{98.48}       & 6.11                 & 98.12                & 8.36                 & 97.54                & 11.94                & 95.35                & 7.67            & 97.37           \\
Deep SVDD~\cite{Deep_SVDD2018}         & 5.71                 & 97.97                & 6.67                 & 97.82                & 9.29                 & 97.57                & 11.81                & 95.14                & 8.37            & 97.13           \\
AnoDDPM~\cite{DDPM2020}         & 7.14                 & 97.29                & 7.33                 & 97.72                & 6.79                 & 97.79                & 12.08                & 95.51                & 8.34            & 97.08           \\
\hline
\textbf{DFG\dag (Ours)} & 5.95                & 97.39               & \textbf{4.00}       & \textbf{98.70}      & 7.07                & 97.63               & 10.28               & 96.17               & 6.83          & 97.47 \\
\textbf{DFG (Ours)} & 5.71                & 98.03               & 5.44                & 98.55               & \textbf{6.71}       & \textbf{98.23}      & \textbf{9.43}       & \textbf{96.62}      & \textbf{6.82} & \textbf{97.86} \\
\hline
\end{tabular}%
}
\end{table*}

\subsection{Different Approaches for Obtaining the Anomalous Cue}
We conduct experiments to compare our DFG model with other commonly used autoencoder-based anomaly detection methods, determining which one extracts more effective anomalous cues. Specifically, we train MemAE~\cite{MemAE2019}, Deep SVDD~\cite{Deep_SVDD2018}, and a DDPM-based anomaly detection method~\cite{DDPM2020} on the same real face dataset. We then use these methods to reconstruct “real” version images on the FAS dataset and calculate the corresponding anomalous cue. For MemAE and Deep SVDD, we directly get the reconstructed “real” version image by feeding the input to the network. 
For DDPM, the image-based diffusion model, we follow the approach presented in AnoDDPM~\cite{AnoDDPM2022} by adding noise to the input image for only 200 steps and then using DDPM to denoise it, thus obtaining the “real” version image corresponding to the input.

Subsequently, we use the anomalous cues obtained by the methods above as the input of our OA-Net, respectively, and conduct the FAS experiments. The training setting is consistent with the Leave-One-Out protocol. We take the ViT-base model without any cue guidance as the baseline in this experiment.
The results in~\Cref{tab:compare_anomaly_detection} show that all methods that introduce the anomalous cues generally exhibit improvement compared to the baseline and the proposed DFG-based anomalous cue guides the FAS model to obtain the best overall performance. Compared with other methods, the DFG method obtains a cue that performs more stable on all the settings, which means our method is more reliable when faced with different scenarios.

We find that the reconstructed background in the anomalous cues may differ from the input, even for real images, which is due to DFG's lack of additional information about the input background. To investigate the role of background in the anomalous cues, we conduct comparative experiments by masking the background regions of the anomalous cues, with the results denoted as ``DFG\dag'' in \Cref{tab:compare_anomaly_detection}. The background mask is obtained by processing the RGB input using the BiSeNet \cite{bisenet_ECCV2018} network. The results show that removing the background regions does not significantly affect the overall experimental outcomes. We attribute this to the fact that the anomalous cues are passed through a trainable cue encoder in the OA-Net, enabling the model to adaptively focus on the information relevant to the presentation attacks while diminishing its emphasis on the noise caused by the background reconstruction.

\subsection{Visualization of the Anomalous Cue}

\begin{figure*}[!ht]%
\flushleft
\includegraphics[width=0.95\textwidth]{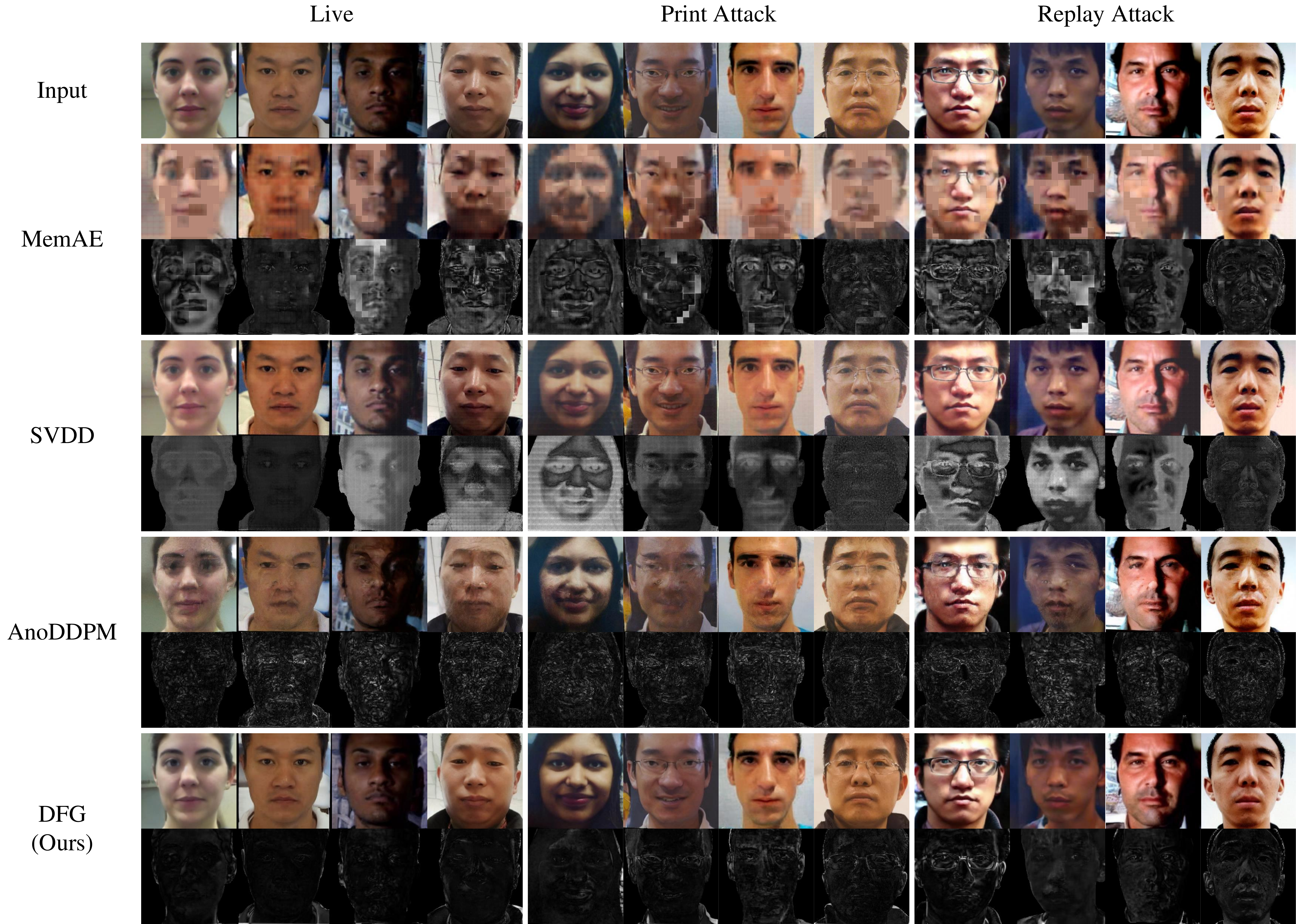}
\caption{Visualization of the generation capabilities for different methods. The top row represents the inputs for each model. Subsequently, the results of each method are presented in two rows: the first row displays the reconstructed images, while the second row illustrates the corresponding anomalous cues.}
\label{fig:reconstruction}
\end{figure*}

The visualization of the reconstruction results obtained by the methods above on the FAS dataset is shown in~\Cref{fig:reconstruction}.
To better highlight the differences in the anomalous cues obtained by each method in the facial region, we present the residual images after masking the background regions.
It can be observed that MemAE and Deep SVDD have relatively poor generating capabilities as the reconstructed images contain obvious generated patterns of noise. This noise present in the generation process can affect the effectiveness of the anomalous cue.
Although the DDPM-based method exhibits relatively good reconstruction results, the reconstructed spoof faces with presentation attacks also closely resemble the input, making the anomalous cues of the real and spoof faces hard to distinguish.

In comparison to the aforementioned methods, the proposed DFG achieves promising results. As for the real images, the reconstructed faces are almost identical to the input. As for the spoof images, the reconstructed faces show a reduction in the artifacts present in the inputs, such as printing patterns and brightness anomalies, which makes the generated images exhibit differences from the inputs. This allows the anomalous cue obtained from DFG to effectively assist the FAS model in distinguishing between real and spoof faces. 

\subsection{Ablation Study}
\subsubsection{Comparisons between Different Structures of OA-Net}
\begin{table*}[t]
\centering
\caption{Comparisons between different structures of OA-Net}\label{tab:compare_FAS_feature_extractor}
\resizebox{\textwidth}{!}{
\begin{tabular}{c|c|c|c|c|c|c|c|c|c|c}
\hline
                & \multicolumn{2}{c|}{\textbf{O\&C\&I   to M}} & \multicolumn{2}{c|}{\textbf{O\&M\&I   to C}} & \multicolumn{2}{c|}{\textbf{O\&C\&M   to I}} & \multicolumn{2}{c|}{\textbf{I\&C\&M   to O}} & \multicolumn{2}{c}{\textbf{Mean}} \\
\cline{2-11}
                & HTER                 & AUC                  & HTER                 & AUC                  & HTER                 & AUC                  & HTER                 & AUC                  & HTER            & AUC             \\
\hline
only image (baseline)          & 6.91                 & 97.06                & 8.00                 & 97.37                & 7.71                 & 97.96                & 16.11                & 92.43                & 9.68          & 96.20          \\
only residual       & 17.14                & 88.44                & 26.44                & 82.42                & 21.57                & 89.00                & 17.36                & 90.92                & 20.63           & 87.69          \\
dual-branch         & 7.14                 & 97.19                & 6.67                 & 98.16                & 11.36                & 95.39                & 13.07                & 94.34                & 9.56            & 96.27          \\
\textbf{OA-Net (Ours)} & \textbf{5.71}        & \textbf{98.03}       & \textbf{5.44}        & \textbf{98.55}       & \textbf{6.71}        & \textbf{98.23}       & \textbf{9.43}        & \textbf{96.62}       & \textbf{6.82}   & \textbf{97.86} \\
\hline
\end{tabular}%
}
\end{table*}
We conduct experiments to study the effectiveness of the obtained anomalous cue and the structure of the OA-Net in our method, and the results are summarized in~\Cref{tab:compare_FAS_feature_extractor}. We still take the ViT-base model without any cue guidance as the baseline in this experiment, denoting the “only image” setting.

The setting “only residual” replaces the image inputs in the setting “only image” with the anomalous cues obtained by the DFG. As observed in the results, the “only residual” setting that solely uses anomalous cues yields inferior performance compared to the “only image” setting that solely uses the original images. The reason is that though the anomalous cue can to some extent indicate where anomaly is present in the input, it cannot fully capture all the attack representations in the original image. The result demonstrates that the anomalous cue should be used together with the original image to assist FAS feature learning.

We then try to adopt the dual-branch architecture to combine the information of the anomalous cue and the original image, denoting the “dual-branch” setting. The dual-branch network uses two ViT-base models to extract the features of the anomalous cue and the original image, respectively, and concatenates the features before feeding them to the softmax classification layer. The result of the “dual-branch” setting is not satisfactory since the FAS feature extractor essentially does not benefit from the anomalous cue via a simple concatenation of the features.

In contrast, our proposed OA-Net leverages the anomalous cue throughout the entire FAS feature extraction process via cross-attention modules, which keeps providing hints that guide the feature extractor to learn from the off-real regions. By effectively utilizing the anomalous cue, our method obtains the best performance on the Leave-One-Out experiments.

\begin{table*}[t]
\centering
\caption{Performance with the anomalous cue generated under different time steps}\label{tab:compare_diffusion_steps}
\resizebox{\textwidth}{!}{
\begin{tabular}{c|c|c|c|c|c|c|c|c|c|c}
\hline
                & \multicolumn{2}{c|}{\textbf{O\&C\&I   to M}} & \multicolumn{2}{c|}{\textbf{O\&M\&I   to C}} & \multicolumn{2}{c|}{\textbf{O\&C\&M   to I}} & \multicolumn{2}{c|}{\textbf{I\&C\&M   to O}} & \multicolumn{2}{c}{\textbf{Mean}} \\
\cline{2-11}
                & HTER                 & AUC                  & HTER                 & AUC                  & HTER                 & AUC                  & HTER                 & AUC                  & HTER            & AUC             \\
\hline
$\hat{t}=1000$         & 5.71                & 97.84               & 7.22                & 97.62               & 11.36               & 95.33               & \textbf{8.19}       & \textbf{97.50}      & 8.12          & 97.07          \\
$\hat{t}=800$      & 5.71                & 98.03               & \textbf{5.44}       & \textbf{98.55}      & \textbf{6.71}       & \textbf{98.23}      & 9.43                & 96.62               & \textbf{6.82} & \textbf{97.86} \\
$\hat{t}=600$      & 5.48                & 98.50               & 6.67                & 98.06               & 8.50                & 97.13               & 8.21                & 97.23               & 7.21          & 97.73          \\
$\hat{t}=400$      & \textbf{4.05}       & \textbf{98.41}      & 7.44                & 97.58               & 7.64                & 97.55               & 10.57               & 95.88               & 7.43          & 97.35          \\
$\hat{t}=200$      & 7.38                & 97.35               & 6.00                & 98.40               & 8.71                & 97.30               & 8.59                & 96.90               & 7.67          & 97.49 \\
\hline
\end{tabular}%
}
\end{table*}

\subsubsection{Effectiveness of the Identity Feature}
\begin{figure*}[!ht]%
\flushleft
\includegraphics[width=0.9\textwidth]{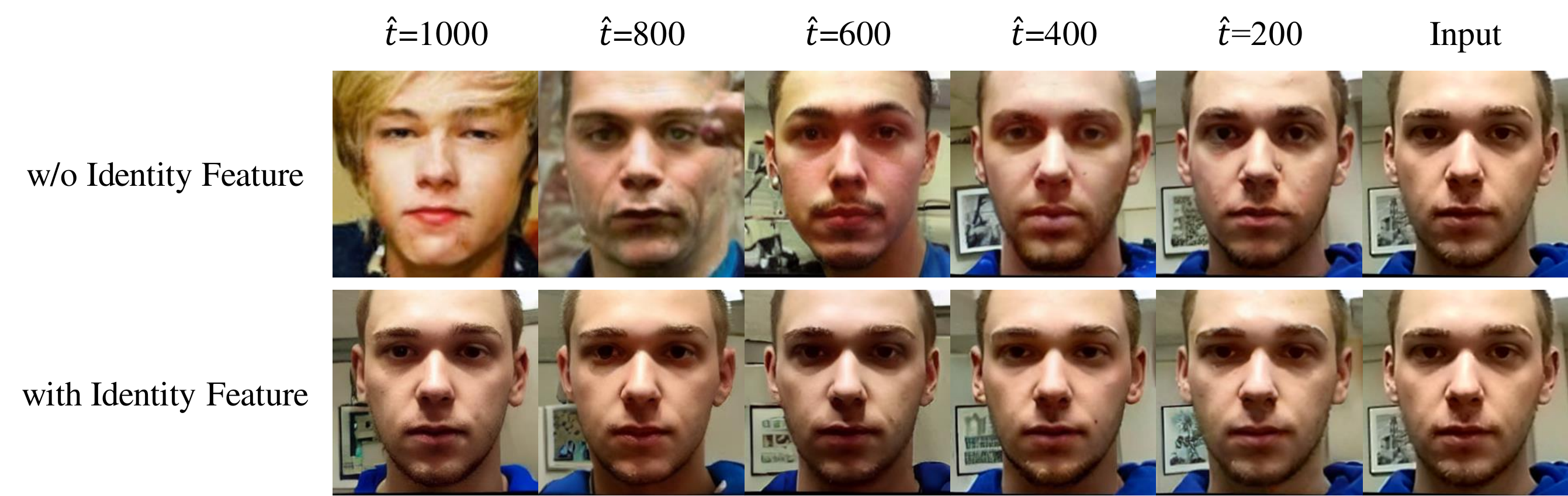}
\caption{Comparison of the DFG trained without/with the identity feature as the conditional guidance, which displays the images reconstructed by the DFG using different diffusion steps $\hat{t}$. The rightmost column represents the corresponding input.}
\label{fig:id_feature}
\end{figure*}
We investigate the influence of the condition identity feature on the training of DFG. We train a DFG model without the identity feature and compare it with a DFG model trained under standard settings. From~\Cref{fig:id_feature}, it can be seen that, without the identity feature as the condition, the DFG tends to exhibit changes in identity while generating the “real” face. In contrast, the DFG model trained with identity guidance keeps the consistency of the identity in the generated face across different diffusion steps. 

Additionally, we find that using only the identity feature ($\hat{t}=1000$) to guide the generation of the “real” face still preserves the pose and expression of the people in the generated images. This suggests that the employed Arcface encoder extracts additional information, such as the pose, beyond the identity feature. 

\subsubsection{Impact of the Diffusion Steps}
\label{sec:compare_diffusion_steps}
We conduct ablation studies on the diffusion steps while generating the anomalous cue, and the results are shown in~\Cref{tab:compare_diffusion_steps}. We find that the model's performance tends to decline when the diffusion steps are either too high or too low. When the diffusion steps are set as $\hat{t}=1000$, the diffusion model reconstructs the image from a latent code in random noise, which introduces more noise into the anomalous cue. However, when the number of the diffusion steps is low, the diffusion model tends to generate images identical to the input, thereby diminishing the attack information in the anomalous cue.

\begin{table*}[t]
\centering
\caption{Performance with the DFG trained on different data quantity}\label{tab:compare_data}
\resizebox{\textwidth}{!}{
\begin{tabular}{c|c|c|c|c|c|c|c|c|c|c}
\hline
                & \multicolumn{2}{c|}{\textbf{O\&C\&I   to M}} & \multicolumn{2}{c|}{\textbf{O\&M\&I   to C}} & \multicolumn{2}{c|}{\textbf{O\&C\&M   to I}} & \multicolumn{2}{c|}{\textbf{I\&C\&M   to O}} & \multicolumn{2}{c}{\textbf{Mean}} \\
\cline{2-11}
                & HTER                 & AUC                  & HTER                 & AUC                  & HTER                 & AUC                  & HTER                 & AUC                  & HTER            & AUC             \\
\hline
25\%      & 6.91                 & 97.71                & \textbf{5.44}        & 98.42                & 8.79                 & 96.46                & 11.96                & 95.03                & 8.27          & 96.91          \\
50\%      & 5.95                 & 97.93                & 6.67                 & 97.78                & 7.14                 & 97.93                & \textbf{9.39}        & \textbf{96.93}       & 7.29          & 97.64          \\
100\%     & \textbf{5.71}        & \textbf{98.03}       & \textbf{5.44}        & \textbf{98.55}       & \textbf{6.71}        & \textbf{98.23}       & 9.43                 & 96.62                & \textbf{6.82} & \textbf{97.86} \\
\hline
\end{tabular}%
}
\end{table*}

\subsubsection{Impact of the Quantity of the Additional Real Faces}
We conduct ablation experiments on the quantity of additional real faces to study its impact on the proposed method. Specifically, we randomly select a certain proportion of the real faces for the training of our DFG and evaluate the performance. The results are shown in~\Cref{tab:compare_data}. It can be observed that with an increase in the data used for DFG, the overall performance of the model improves. This suggests that when we provide more data of real faces during training, the proposed DFG can actually gain better knowledge of what a real face should be like and thus generate more effective anomalous cues.
The result above indicates there is potential for future improvement in the performance of AG-FAS if more real faces can be obtained.

\subsection{Generalization to Other Datasets}

\begin{table*}[ht]
\centering
\caption{A Larger-scale FAS cross-domain benchmark}\label{tab:large_benchmark}
\resizebox{\textwidth}{!}{
\begin{tabular}{c|c|c|c|c|c|c|c|c|c|c}
\hline
                & \multicolumn{2}{c|}{\textbf{HKBU-MARs}} & \multicolumn{2}{c|}{\textbf{WFFD}} & \multicolumn{2}{c|}{\textbf{Rose-Youtu}} & \multicolumn{2}{c|}{\textbf{SiW}}& \multicolumn{2}{c}{\textbf{WMCA}}   \\
\cline{2-11}
                & HTER                & AUC                 & HTER               & AUC               & HTER            & AUC             & HTER               & AUC                & HTER           & AUC             \\
\hline
SSDG-R~\cite{SSDG_CVPR2020}          & 23.53              & 84.22             & 37.74           & 65.29           & 12.22              & 94.54              & 4.97          & 98.68          & 14.61         & 91.98                   \\
SSAN-R~\cite{SSAN_CVPR2022}          & 22.12              & 85.68             & 40.24           & 63.49           & 14.15              & 92.80              & 8.85          & 97.01          & 16.79         & 90.40                   \\
SA-FAS~\cite{SA-FAS2023}           & 31.63              & 74.86             & 41.34           & 60.67           & 16.42              & 90.33              & 6.79          & 98.01          & 18.91         & 88.08                   \\
\textbf{AG-FAS (Ours)} & \textbf{17.98}     & \textbf{90.43}    & \textbf{35.44}  & \textbf{67.23}  & \textbf{11.77}     & \textbf{95.13}     & \textbf{3.18} & \textbf{99.50} & \textbf{9.95} & \textbf{96.84}    \\
\hline
\end{tabular}%
}
\end{table*}

To further validate the effectiveness of AG-FAS, we conduct experiments on a broader range of commonly used datasets in the field of FAS to test the generalization ability of our approach. Specifically, we train different DG-based models on MSU-MFSD, CASIA-FASD, Idiap Replay-Attack, and OULU-NPU and then evaluate the performance of these models on other commonly used FAS datasets, i.e., HKBU-MARs, WFFD, Rose-Youtu, SiW, and WMCA, respectively. The experimental results are presented in~\Cref{tab:large_benchmark}. As can be seen, our method achieves the best performance on all the testing FAS datasets, further demonstrating that anomalous cue guides the model to catch more generalized attack representations in the face images.

It is worth noting that among the FAS datasets used for testing, the HKBU, WFFD, and WMCA datasets include 3D mask and wax figure presentation attacks, respectively, which are not contained in the four datasets used for training. The experimental results on these two datasets prove that our method also performs well in the challenging FAS task with unknown presentation attacks.

\section{Conclusion}
In this paper, we propose an Anomalous cue Guided Face Anti-Spoofing (AG-FAS) method, which can effectively leverage additional data of real faces for improving model generalization on cross-domain FAS tasks. 
We train a De-fake Face Generator (DFG) on a large-scale real face only dataset, enabling it to generate the “real” version of any input face. The residual of the generated “real” face and the input reflects the off-real region and serves as the anomalous cue for FAS. Specifically, we construct an Off-real Attention Network (OA-Net) that utilizes the anomalous cue to allocate its attention toward the off-real region in the input, achieving a more generalized FAS feature. We demonstrate the effectiveness of our method through both theoretical analysis and extensive cross-domain FAS experiments with unseen scenarios and unknown presentation attacks.

As this work solely relies on two additional datasets with 130K real faces to obtain the anomalous cue, one of our future directions will consider training the DFG on larger scale face datasets from the Internet, like Glint360K~\cite{glint360k2022}, to explore the possibility of further improving generalization.


\normalem
\bibliographystyle{IEEEtran}
\bibliography{egbib}

\newpage

 

\begin{IEEEbiography}[{\includegraphics[width=1in,height=1.25in,clip,keepaspectratio]{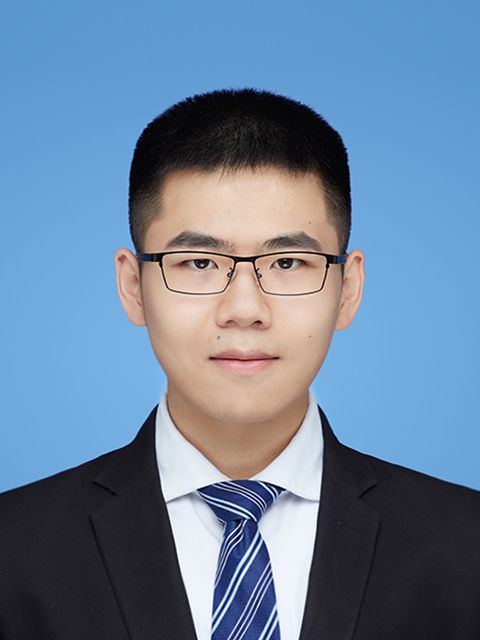}}]{Xingming Long} (Student Member, IEEE) received the B.S. degree in computer science from Tsinghua University in 2021. He is currently pursuing the M.S. degree with the Institute of Computing Technology (ICT), Chinese Academy of Sciences (CAS). His research interests include face anti-spoofing and domain generalization.
\end{IEEEbiography}

\begin{IEEEbiography}[{\includegraphics[width=1in,height=1.25in,clip,keepaspectratio]{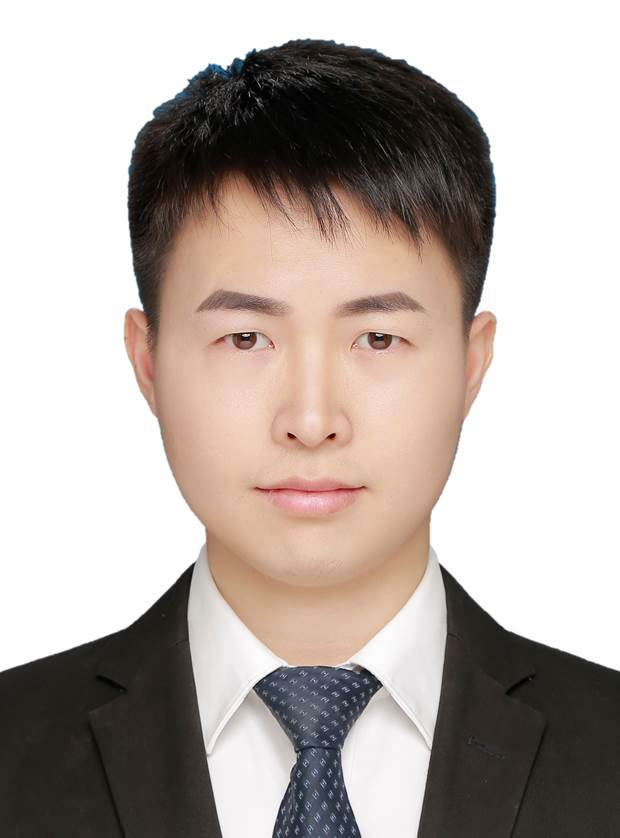}}]{Jie Zhang}
(Member, IEEE) received the Ph.D. degree from the University of Chinese Academy of Sciences, Beijing, China. He is currently an Associate Professor with the Institute of Computing Technology (ICT), Chinese Academy of Sciences (CAS). His research interests include computer vision, pattern recognition, machine learning, particularly include face recognition, image segmentation, weakly/semi-supervised learning, and domain generalization.
\end{IEEEbiography}

\begin{IEEEbiography}[{\includegraphics[width=1in,height=1.25in,clip,keepaspectratio]{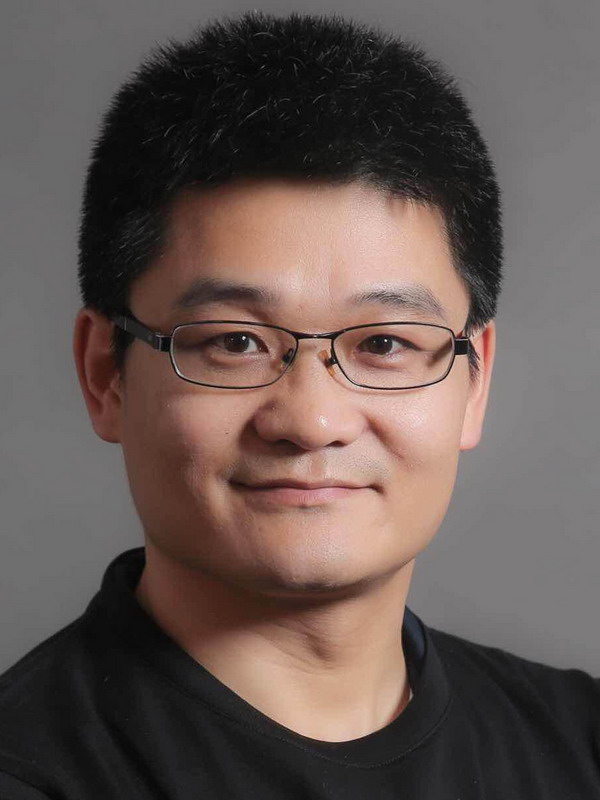}}]{Shiguang Shan}
(Fellow, IEEE) received the Ph.D. degree in computer science from the Institute of Computing Technology (ICT), Chinese Academy of Sciences (CAS), Beijing, China, in 2004. Since 2010, he has been a Full Professor with ICT, CAS, where he is currently the Director of the Key Laboratory of Intelligent Information Processing. His research interests include computer vision, pattern recognition, and machine learning. He has published more than 300 articles in related areas. He was a recipient of the China’s State Natural Science Award in 2015 and the China’s State S\&T Progress Award in 2005 for his research work. He served as an Area Chair for many international conferences, including CVPR, ICCV, AAAI, IJCAI, ACCV, ICPR, and FG. He was/is an Associate Editor of several journals, including IEEE TRANSACTIONS ON IMAGE PROCESSING, Neurocomputing, CVIU, and PRL.
\end{IEEEbiography}

\vfill

\end{document}